\documentclass[]{fairmeta}

\usepackage{makecell}
\usepackage{wrapfig}
\usepackage{tabularx}
\usepackage{textcomp}
\IfFileExists{stfloats.sty}{\usepackage{stfloats}}{}
\usepackage{url}
\usepackage{verbatim}
\usepackage{titlesec}
\usepackage{adjustbox}
\usepackage{multirow}
\usepackage[sc]{mathpazo}
\usepackage{comment}
\usepackage{amsmath,amssymb}
\usepackage{colortbl}
\usepackage[numbers]{natbib}
\usepackage{color}
\usepackage{booktabs}
\usepackage{graphicx}
\usepackage{subcaption}
\RequirePackage{xspace}
\makeatletter
\DeclareRobustCommand\onedot{\futurelet\@let@token\@onedot}
\def\@onedot{\ifx\@let@token.\else.\null\fi\xspace}

\makeatother
\usepackage{xcolor}
\usepackage{array}
\usepackage{caption}
\usepackage{float}
\renewcommand{\paragraph}[1]{\vspace{1.25mm}\noindent\textbf{#1}}
\usepackage{enumitem}
\usepackage{fvextra}            
\newcolumntype{Y}{>{\raggedright\arraybackslash}X}

\definecolor{LoneFrame}{HTML}{8C6BB1}
\definecolor{LoneBg}{HTML}{F4F0FA}
\definecolor{LtwoFrame}{HTML}{2B8CBE}
\definecolor{LtwoBg}{HTML}{EEF7FB}
\definecolor{LthreeFrame}{HTML}{31A354}
\definecolor{LthreeBg}{HTML}{F0F8F0}
\definecolor{LfourFrame}{HTML}{F16913}
\definecolor{LfourBg}{HTML}{FFF3E8}
\definecolor{LfiveFrame}{HTML}{D94801}
\definecolor{LfiveBg}{HTML}{FFF0E6}

\newcommand{\PromptFileBox}[4]{%
  \par\nobreak\vspace{1.5pt}%
  \noindent\colorbox{#2}{\textcolor{#1}{$\blacksquare$}\,{\scriptsize\textbf{#3}}}%
  \par\nobreak\vspace{0.5pt}%
  \VerbatimInput[fontsize=\scriptsize,baselinestretch=0.92,breaklines=true,breakanywhere=true,breaksymbolleft={},breaksymbolright={},tabsize=2]{#4}%
  \vspace{1.5pt}%
}
\newcommand{\LoneFile}[2]{\PromptFileBox{LoneFrame}{LoneBg}{$L_1$ --- #1}{#2}}
\newcommand{\LtwoFile}[2]{\PromptFileBox{LtwoFrame}{LtwoBg}{$L_2$ --- #1}{#2}}
\newcommand{\LthreeFile}[2]{\PromptFileBox{LthreeFrame}{LthreeBg}{$L_3$ --- #1}{#2}}
\newcommand{\LfourFile}[2]{\PromptFileBox{LfourFrame}{LfourBg}{$L_4$ --- #1}{#2}}
\newcommand{\LfiveFile}[2]{\PromptFileBox{LfiveFrame}{LfiveBg}{$L_5$ --- #1}{#2}}

\DeclareUnicodeCharacter{2192}{$\rightarrow$}
\DeclareUnicodeCharacter{21B3}{$\hookrightarrow$}
\DeclareUnicodeCharacter{00D7}{$\times$}
\DeclareUnicodeCharacter{2248}{$\approx$}
\DeclareUnicodeCharacter{2014}{---}
\DeclareUnicodeCharacter{2013}{--}
\DeclareUnicodeCharacter{2011}{-}
\DeclareUnicodeCharacter{2019}{'}

\newcommand{\stspad}{Slay the Spire 2\xspace}
\newcommand{\yesmark}{\textcolor{green!45!black}{\boldmath$\checkmark$\unboldmath}}
\renewcommand{\partmark}{\textcolor{orange!85!black}{$\blacktriangle$}}
\newcommand{\nomark}{\textcolor{gray}{\textbf{--}}}

\title{AgenticSTS: A Bounded-Memory Testbed for Long-Horizon LLM Agents}

\author[1,2]{Xiangchen Cheng}
\author[2,3]{Yunwei Jiang}
\author[1,3,4]{Jianwen Sun}
\author[1,3,4]{Zizhen Li}
\author[1]{Chuanhao Li}
\author[1]{Xiangcheng Cao}
\author[1]{Yihao Liu}
\author[3,5]{Fanrui Zhang}
\author[2]{Li Jin}
\author[1,\dagger]{Kaipeng Zhang}

\affiliation[1]{Alaya Lab}
\affiliation[2]{Shanghai Jiao Tong University}
\affiliation[3]{Shanghai Innovation Institute}
\affiliation[4]{Nankai University}
\affiliation[5]{University of Science and Technology of China}

\contribution[\dagger]{Corresponding author}

\abstract{

Memory for a long-horizon LLM agent is a contract about what each future decision is allowed to see. The simplest contract appends past observations, tool calls, and reflections to every prompt, which makes prior context easy to access but also turns it into a jumbled mixture in which the effect of any single memory component is hard to isolate. We introduce and instrument an alternative bounded contract: every decision is made from a fresh user message assembled by typed retrieval, with no raw cross-decision transcript appended. The prompt thus stays bounded across runs of any length, and any single layer can be ablated in isolation. We instantiate the contract in \stspad{}, a closed-rule stochastic deck-building game whose runs require hundreds of tactical and strategic decisions. A public online benchmark of frontier LLMs on the same game reports zero wins at the lowest difficulty across five configurations, and the developer-reported human win rate at the same difficulty is $16\%$; the task is hard but not saturated. Within our harness, a fixed-$A_0$ ablation shows the largest observed difference when triggered strategic skills are enabled: the no-store baseline wins $3/10$ games and adding the skill layer $6/10$. At this sample size the comparison is directional rather than statistically decisive (Fisher exact $p\approx0.37$); a cross-backbone probe and public accumulating-context baselines are reported as operational comparisons rather than controlled tests of the contract variable itself. We release a reproducible testbed: $298$ completed trajectories with condition tags, frozen memory/skill snapshots, prompt records, and analysis scripts---an agent design and a validated, reusable methodology for studying how explicit memory layers shape long-horizon LLM-agent decisions.
}

\projectpage{\url{https://github.com/AlayaLab/AgenticSTS}}
\github{\url{https://github.com/AlayaLab/AgenticSTS}}
\data{\url{https://huggingface.co/datasets/ShandaAI/AgenticSTS-trajectories}}
\correspondence{\email{kaipeng.zhang@shanda.com}}
\date{\today}

\begin{document}
\maketitle

\section{Introduction}
\label{sec:intro}

\begin{figure}[t]
\centering
\includegraphics[width=\linewidth]{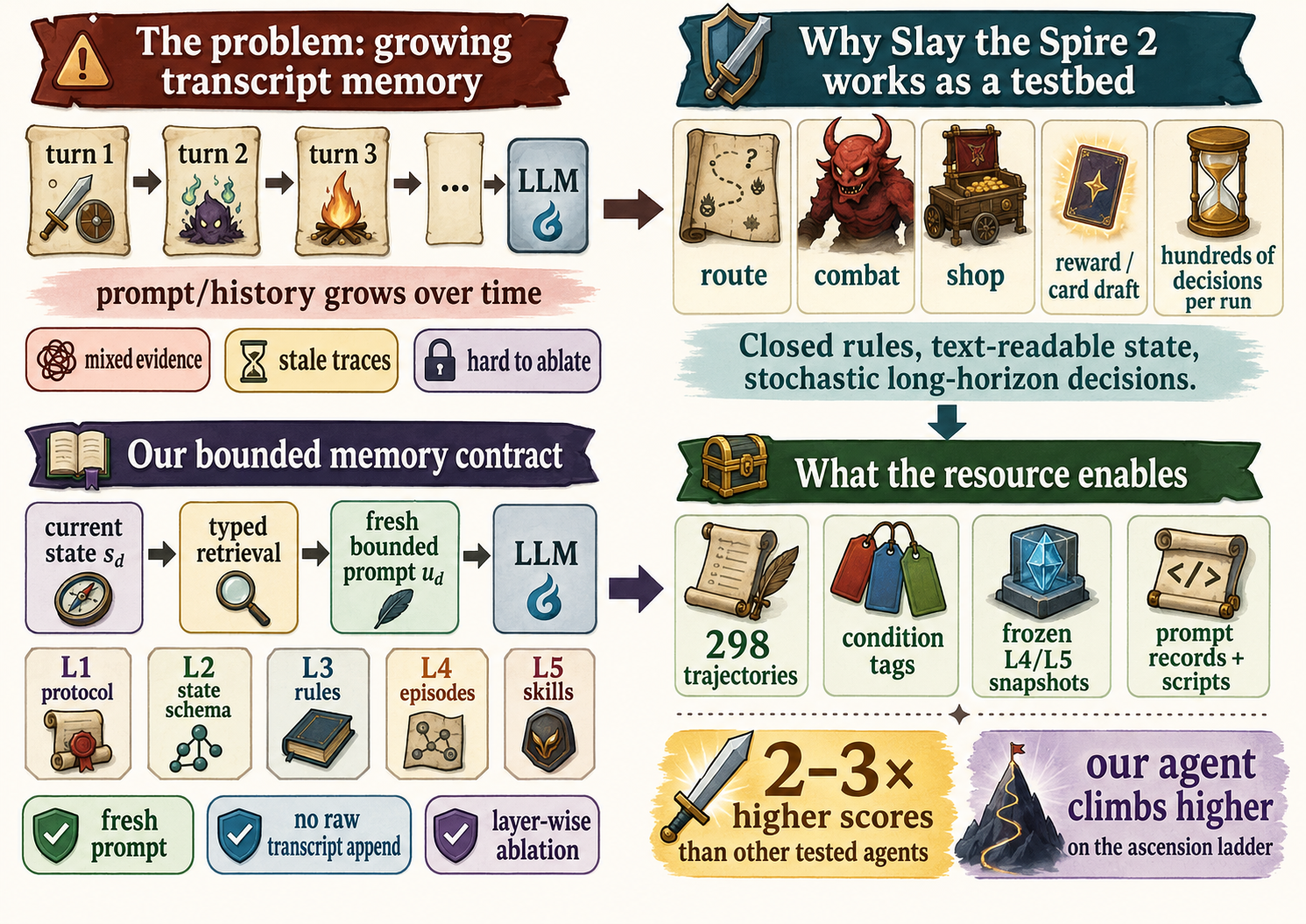}
\caption{Overview of our paper: a bounded, typed memory contract
turns long-horizon LLM-agent memory into an ablatable evaluation
surface. Summary performance labels in this schematic (e.g.\
relative scores and ladder reach) are illustrative; the exact
numbers, denominators, and caveats---including that cross-agent
comparisons are operational rather than matched ablations and that
win-rate differences are directional at our sample size---are given
in \S\ref{sec:results}--\ref{sec:competitors}.}
\label{fig:overview-pro}
\end{figure}

\begin{figure*}[t]
\centering
\includegraphics[width=\linewidth]{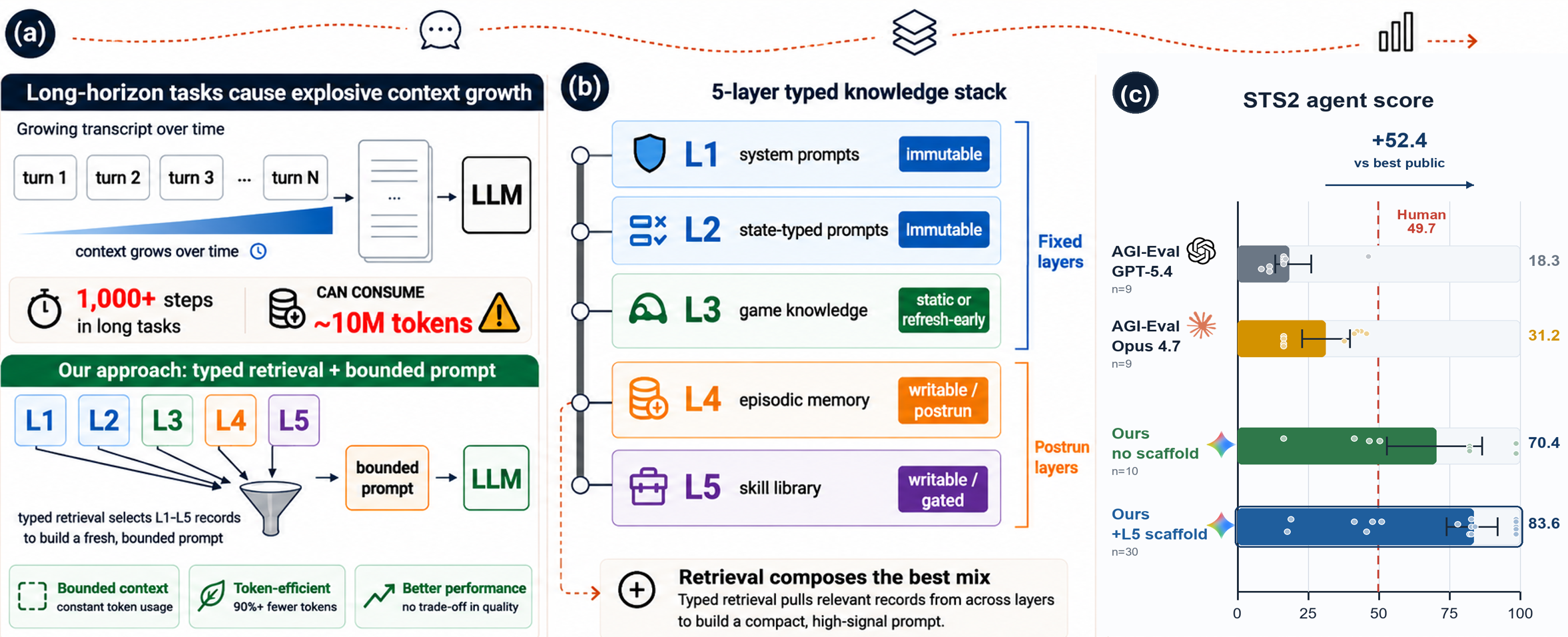}
\caption{Typed retrieval as a bounded-memory contract:
\textbf{(a)}~per-decision composition, \textbf{(b)}~the five typed
layers, and \textbf{(c)}~AgenticSTS scores vs AGI-Eval rows
and the community human reference
\citep{sts2fun-community-stats-2026}. In-panel summary labels (e.g.\
``no trade-off'', token-efficiency and score-gap figures) are
illustrative; cross-backbone and external comparisons are operational
context, not matched ablations of the memory contract, and
within-harness win-rate differences are directional at our sample
size---see \S\ref{sec:results}--\ref{sec:competitors} for exact
numbers and caveats.}
\label{fig:overview}
\end{figure*}

For a long-horizon LLM agent, memory is not a place to store
text; it is a contract about what each future decision is allowed
to see. One common contract appends past observations, tool calls,
and reflections to the next prompt
\citep{react2023,reflexion2023,voyager2023}.
Another distills prior experience into typed records and retrieves
only the pieces selected for the current decision
\citep{coala-2024,memgpt2024,a-mem-2025,memoryos2025,generative-agents-2023}. This choice
is not just an engineering detail: it determines what evidence the
model sees, what stale information can re-enter a decision, and
which component can be ablated when the agent succeeds or fails.
We therefore ask whether a long-horizon benchmark
\citep{sinha-horizon-2026,kwa-metr-2025} can make the memory
interface bounded, inspectable, and reusable rather than treating
context growth as an implicit default. Practitioners increasingly
frame this through the \emph{agent loop} and its context or ``loop''
engineering~\citep{anthropic-context-2025,memory-survey-2025}---the
per-iteration choice of which prior experience to place in a finite
window; our bounded contract is a formal, ablatable answer to the
memory stage of that loop.

We instantiate this question in \stspad{}, a roguelike
deck-building game. A run is a stochastic strategic campaign:
the agent chooses map routes, fights turn-based battles, drafts
cards, buys or skips items, and preserves scarce health over
many delayed consequences. The rule space is closed, symbolic, and
text-readable, so game facts and legal actions can be supplied as
structured records rather than pixels. At the same time, success
requires hundreds of local and long-range decisions under random
card draws, rewards, enemies, events, and difficulty modifiers.
Public benchmarks indicate that the task is not already saturated.
The game ships an ascension difficulty system that runs from $A_0$
(the easiest tier) up to $A_{10}$ (the hardest). At $A_0$,
a publicly available benchmark of frontier LLMs on
this game reports zero wins across five model configurations, and
the developer-reported human win rate is $16\%$
\citep{agi-eval-spire-2026,megacrit-neowsletter-may2026}. Together these numbers make \stspad{} a hard but unsaturated
testbed for studying long-horizon LLM-agent memory.

Our agent, AgenticSTS, implements typed retrieval as a bounded
memory contract.
For every decision, the user message is freshly composed from five
slots: fixed protocol instructions ($L_1$), state-specific schemas
and legal action formats ($L_2$), retrieved game rules ($L_3$),
episodic summaries ($L_4$), and triggered strategic skills ($L_5$).
The slots differ in mutability and experimental role: $L_1$ and
$L_2$ are fixed; $L_3$ can be filtered; $L_4$ and $L_5$ can be
disabled, frozen, or made writable between runs through postrun
analysis; raw
cross-decision transcripts are not appended. The resulting prompt
interface turns memory from ``how much history fits'' into ``which
typed evidence is selected,'' a form that can be inspected and
re-aggregated across conditions.

Across $298$ completed trajectories, we run two main evidence
streams and a cross-backbone probe.
The first holds difficulty fixed at $A_0$ and varies the five
memory slots: a no-scaffold baseline wins $3/10$ games (Wilson
95\% confidence interval $[10.8\%, 60.3\%]$, i.e., the true
win-rate is likely between $10.8\%$ and $60.3\%$), and the largest
observed difference coincides with enabling $L_5$ skills, which
reach $6/10$ in each scaffolded cell ($[31.3\%, 83.2\%]$). At this
sample size the difference is directional rather than statistically
decisive (Fisher exact $p\approx0.37$), and the five typed slots
remain individually ablatable. The second stream climbs the
difficulty ladder with $L_4$ and $L_5$ stores that update
between runs, reaching high-difficulty probes at $A_6$--$A_8$.
A cross-backbone probe (the underlying LLM is swapped among
Gemini, Qwen, and DeepSeek) exercises the same ablation surface on
other model families; the frozen stack is backbone-sensitive, with
model-specific caveats detailed in \S\ref{sec:results:modeb}.

The released archive supports re-aggregation and cross-condition
re-analysis: a reader can recompute the headline fixed-$A_0$ cells
or slice trajectories by condition tag using the included scripts.
We ship the condition-tagged trajectories, the SHA-anchored
$L_4{+}L_5$ snapshots used in the fixed-$A_0$ matrix, decision-time
prompt records, and Wilson/bootstrap analysis scripts.

\paragraph{Contributions.}
We contribute (see Figure~\ref{fig:overview-pro}):
\begin{enumerate}[leftmargin=1.5em,labelsep=0.3em,itemsep=0pt,topsep=0pt,partopsep=0pt,parsep=0pt,label=(\roman*)]
\item a per-decision composition interface assembling prompts
from typed $L_1$--$L_5$ slices rather than a raw transcript;
\item evidence that, under the bounded contract, the largest
observed $A_0$ difference coincides with enabling triggered $L_5$
skills ($3/10\to6/10$, directional at this sample size rather than
statistically significant), that the typed slots remain individually
ablatable across three model backbones, and that the surface admits
ladder probes at high-difficulty $A_6$--$A_8$;
\item a reusable archive of $298$ trajectories with condition
tags, SHA-anchored $L_4/L_5$ snapshots, prompts, and
Wilson/bootstrap scripts for community study of context use in
long-horizon agents.
\end{enumerate}
Claims concern layer separability inside the contract; a matched
accumulating-context comparison is future work (Limitations).
Prior prompt-history \citep{react2023,reflexion2023,voyager2023},
structured-memory \citep{memgpt2024,mem0-2025,memoryos2025,gam-2026},
and skill-library agents \citep{skillsbench2026,skillos2026,%
memento-skills-2026} motivate the design space.
\S\ref{sec:testbed} presents the testbed; \S\ref{sec:arch} the
contract; \S\ref{sec:methodology} the protocol;
\S\ref{sec:results} the results.

\section{Related Work}
\label{sec:related}

We build on four recent research threads---prompt-history agents,
externalized memory, skill libraries, and long-horizon game
testbeds---and target their joint problem: which slice of prior
experience enters each decision, and whether that route can itself
be ablated.

\paragraph{Loop and context engineering.}
In practice, agent design is increasingly framed as \emph{loop
engineering}: specifying a goal, tools, termination, and---centrally---the
memory and context policy of a control loop that runs over hundreds of
steps and multiple sessions~\citep{anthropic-context-2025}. The recurring
failure mode is context growth: appending full transcripts and tool logs
to every call overflows the window and dilutes attention, whereas
``token-poor'' loops keep only the last few messages or short summaries
and therefore depend on an explicit memory store~\citep{memory-survey-2025}.
This is exactly the axis our contract isolates. That literature is largely
qualitative and centered on coding agents; the academic threads below
supply the mechanisms, and we make the memory stage of the loop a typed,
bounded, ablatable contract on a hard long-horizon game.

\paragraph{Long-horizon LLM agents and prompt-visible history.}
ReAct and Reflexion made it natural for observations and
self-critiques to reappear in later LLM
calls \citep{react2023,reflexion2023}. Recent horizon analyses
\citep{sinha-horizon-2026,wang-planning-2026,horizon-mirage-2026,%
kwa-metr-2025} examine how small errors compound over many steps.

\paragraph{Typed and structured memory.}
MemGPT, Mem0, MemoryOS, GAM, hierarchical procedural memory, and
Agent Workflow Memory \citep{memgpt2024,mem0-2025,memoryos2025,%
gam-2026,macla-2025,awm-2025} move information out of raw message
history into external stores. Adjacent typed-memory frameworks
\citep{coala-2024,generative-agents-2023,a-mem-2025} group memory
by capacity; we instead role-type slots by mutability and retrieval
source. Most evaluation is in dialogue or QA; in our setting
retrieval feeds an action policy in a stochastic environment.

\paragraph{Self-evolving skill libraries.}
Voyager pioneered an external library of agent-written skills
\citep{voyager2023}; SkillsBench, SkillOS, Memento-Skills,
SAGE/SkillRL, SkillWeaver, ExpEL, and DyStIL extend the design
\citep{skillsbench2026,skillos2026,memento-skills-2026,%
sage-skillrl-2025,skillweaver2025,expel2024,dystil2025}. In the
SoK notation $S=(C,\pi,T,R)$ \citep{sok-agentic-skills-2026}, our
$L_5$ guides correspond to $(C,\pi)$: a trigger selects a prose
policy for the next decision.

\paragraph{LLM agents on games.}
Games such as Crafter, NetHack, BALROG, LMGame-Bench, DSGBench,
Gameverse, and RAGEN \citep{crafter2022,nethack2020,balrog2024,lmgame-bench-2025,%
dsgbench-2025,gameverse2026,ragen2025} provide stochastic testbeds
for agents. Card-game work includes end-to-end policy networks
\citep{xiao-hearthstone-2023}, LoRA-tuned draft models
\citep{urzagpt-2025}, cross-card-game LLM evaluation
\citep{wang-cardgames-2025}, and LLM play on the original Slay the
Spire with simplified rule sets
\citep{bateni-whitehead-2024,rethinking-agent-design-2025}.

\paragraph{Public \stspad{} LLM agents.}
The public \stspad{} ecosystem---STS2MCP, HermesBridge, AI-Spire,
CharTyr \citep{sts2mcp-2026,hermesbridge-2026,ai-spire-2026,%
chartyr-sts2-2026}---does not report a matched ablation over
prompt strictness, episodic memory, and triggered skills.
AGI-Eval~\cite{agi-eval-spire-2026} lists zero $A_0$ victories across
five frontier-model rows with mixed denominators.

\begin{table*}[t]
\centering\small
\caption{Positioning map for prior-experience interfaces. Green checkmarks mark a central axis; orange triangles mark a partial or optional axis. Citations appear in the surrounding paragraphs.}
\label{tab:memcompare}
\setlength{\tabcolsep}{6pt}
\renewcommand{\arraystretch}{1.08}
\begin{tabularx}{\textwidth}{@{}p{0.22\textwidth}cccccY@{}}
\toprule
Family & Transcript & Typed memory & Skills & Layer ablation & Game policy & Gap \\
\midrule
Prompt-history / replay & \yesmark & \nomark & \partmark & \nomark & \partmark & attribution \\
Structured memory       & \nomark & \yesmark & \nomark & \partmark & \nomark & policy \\
Skill-library agents    & \partmark & \partmark & \yesmark & \partmark & \partmark & contract \\
\textbf{Ours}            & \textbf{\nomark} & \textbf{\yesmark} & \textbf{\yesmark} & \textbf{\yesmark} & \textbf{\yesmark} & \textbf{joint test} \\
\bottomrule
\end{tabularx}
\end{table*}

We combine these threads in one bounded-memory contract, so future
work can compare alternative contracts under the same harness.

\section{The \stspad{} Testbed}
\label{sec:testbed}

\stspad{} is a turn-based deck-building roguelike: an agent builds
a deck during a run, fights stochastic battles, chooses routes and
rewards, and climbs an ordinal difficulty ladder. The game is
useful for evaluating LLM-agent memory because it is long-horizon
but not visually opaque. Rules, cards, relics, enemies, events,
legal actions, and state transitions can be represented as text
records, while success still requires sustained planning across
many contingent decisions. We release the resulting runs as a
reusable evaluation resource (\S\ref{sec:methodology:repro}).

\subsection{Four properties of \stspad{} as an LLM testbed}
\label{sec:testbed:props}

\textbf{(\emph{P1}) Closed, enumerable, LLM-readable rule space.}
Public database snapshots index hundreds of typed records (576
cards, 293 relics, 115 monsters, 87 encounters, and 66 events in
Spire Codex's May~2026 API; these are database counts, not unique
experimental-patch counts) \citep{spirecodex-2026}. Unlike
pixel-rendered Crafter \citep{crafter2022} or the inherited code
complexity of NetHack \citep{nethack2020}, \stspad{} has a compact
rule space that can be loaded into a typed knowledge layer
(\S\ref{sec:arch}). This makes $L_3$ part of the evaluation
substrate rather than a hidden source of game knowledge.

\textbf{(\emph{P2}) Empirically long horizon.} A typical run lasts
a median $\sim$80~min wall-clock (IQR 37--109\,min) and contains
67 LLM strategic calls (IQR 27--105). Roughly $500$ additional
per-run decisions, such as combat targets, treasure choices, map
nodes, and hand selection, are mechanically resolved or routed to a
fast tier with bounded combat context (\S\ref{sec:arch}). This is
precisely the regime where message-history accumulation becomes
costly relative to typed recomposition
\citep{sinha-horizon-2026,horizon-mirage-2026,kwa-metr-2025}.

\textbf{(\emph{P3}) Multi-axis stochasticity.} Random card draw,
shuffle order, reward offerings, map paths, relic effects, elite
and event placements, and Ascension modifiers prevent simple
trajectory replay. A strong policy must generalize over states, not
memorize a fixed route.

\textbf{(\emph{P4}) State-conditioned combat math.} Damage and
status pipelines combine hand contents, enemy intent, block timing,
and effects such as \emph{vulnerable, weak, strength}, and
\emph{dexterity}. The agent must compute from the current state;
web-like recall is much less useful than state-conditioned
calculation.

\subsection{Ascension ladder and scoring}
\label{sec:testbed:difficulty}

The 11-level Ascension ladder ($A_0$--$A_{10}$, with $A_{10}$ the
maximum) gives the testbed an ordinal difficulty scale. Higher
Ascensions stack modifiers that change strategic priorities, so
climbing the ladder is not just repeating $A_0$ with larger
numbers. We therefore use two complementary evidence streams: a
fixed-$A_0$ matrix that isolates components at one difficulty
(\S\ref{sec:methodology}) and an auto-mode ladder in which the
agent advances after victories and retries after defeats
(\S\ref{sec:methodology}).

Runs are scored with a derived analysis score:
\begin{equation}
s =
\begin{cases}
100 & \text{if victory,}\\
\mathrm{floor} + \tfrac{52}{3}\cdot\mathrm{bosses} & \text{otherwise,}
\end{cases}
\label{eq:score}
\end{equation}
where $\mathrm{bosses}$ counts cleared act bosses (0/1/2 by reached
floor for non-victory runs, 3 for victories; full mapping in
Appendix~\ref{app:configs}). The value is recomputed from outcome,
floor, and boss-count fields, not copied from the raw archive
\texttt{score} field. The $52/3$ coefficient calibrates three
cleared bosses to $52$ points, the approximate mid-Act-3 floor
reach. A $\pm 10\%$ perturbation of the coefficient checks
score-based qualitative comparisons; win-rate claims do not depend
on this score scale.

\subsection{Data corpus, release, and harness}
\label{sec:testbed:data}

The released archive contains $298$ completed independent game
trajectories spanning fixed-$A_0$ ablations, cross-backbone probes,
and auto-mode ascension runs. Each trajectory records target and
reached Ascension, outcome, wall-clock duration, LLM-call counts,
condition tag, and the active memory/scaffold setting. The headline
fixed-$A_0$ comparison uses a pre-specified balanced subset of 50
completed games, namely the first ten per condition under the
frozen configuration. Other completed trajectories support the
cross-backbone and ladder diagnostics rather than entering the
fixed-$A_0$ estimate. Frozen $L_4{+}L_5$ stores and per-condition tags
are released with the public artifact archive
\citep{deal-checklist-2025,liu-versioning-2025}.

The game alone is not the full benchmark: a game interface must
be paired with a decision protocol. Public \stspad{}
implementations \citep{sts2mcp-2026,hermesbridge-2026,%
ai-spire-2026,chartyr-sts2-2026} let LLMs act in the game, while
the independently evaluated AGI-Eval configurations
\citep{agi-eval-spire-2026} report no listed victory. Our
architecture (\S\ref{sec:arch}) supplies the bounded-memory
contract that makes the released trajectories a reusable evaluation
surface.

\section{Architecture: Per-Decision Typed Retrieval}
\label{sec:arch}

This section specifies what the LLM is allowed to see when it makes
a move. The agent never appends the raw message turns from earlier
decisions. Instead, it rebuilds each decision prompt from five typed
knowledge layers (Figure~\ref{fig:overview}b). Any information that
survives across decisions must first be written into a bounded store;
in our experiments, postrun extraction and skill discovery can write
back only to $L_4/L_5$.

The contract gives the resource four evaluation handles that a raw
prompt-history setup usually hides: horizon growth is capped by
slot budgets; retrieved evidence is labeled by layer; $L_4$ and
$L_5$ can be toggled without rewriting the whole prompt; and runs,
stores, prompts, and scripts carry condition tags for reuse.

\subsection{Per-decision compositional context}
\label{sec:arch:bounded}

In ReAct/Reflexion-style agents \citep{react2023,reflexion2023}, the
model may see a growing log of earlier states, tool calls, and
self-critiques. AgenticSTS uses a different interface: for
decision $d$ at state $s_d$, the engine retrieves from
$L_1,\ldots,L_5$ and composes a fresh user message
\begin{equation}
u_d = \pi\bigl(L_1, L_2(s_d), L_3(s_d), L_4(s_d), L_5(s_d)\bigr),
\label{eq:composition}
\end{equation}
sent to the LLM as $\langle\mathrm{sys},u_d\rangle$. The design still
allows bounded typed summaries, a per-run Strategic Thread, and
same-decision repair retries. What it disallows is an unbounded
transcript that grows because the run has been long.

\begin{figure*}[t]
\centering
\includegraphics[width=0.96\textwidth]{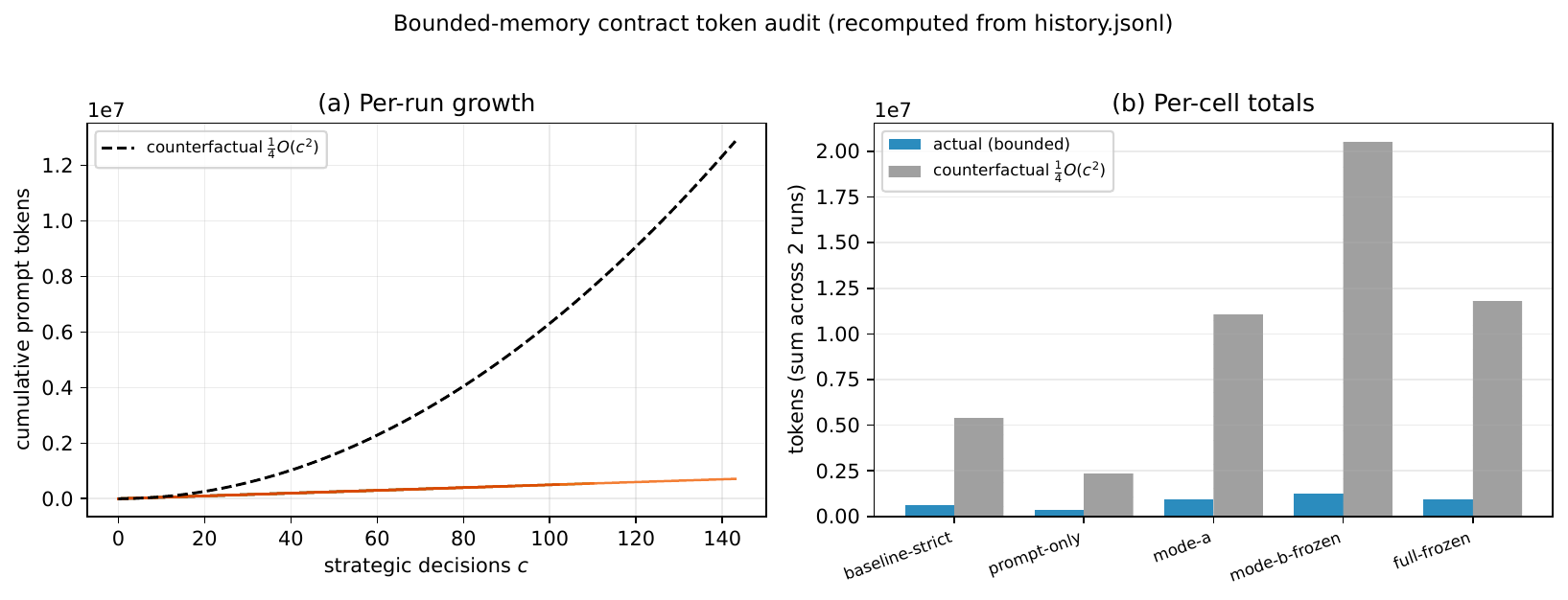}
\caption{Token audit of the bounded-memory contract on ten
fixed-$A_0$ runs (two per cell); dashed line is a transcript-appending
counterfactual at $\tfrac{1}{4}$ of naive $O(c^2)$ growth (median
tokens/call $\times c(c{+}1)/8$, a moderate prompt-caching discount).
The audit illustrates context-growth mechanics under each contract;
at two runs per cell it is not a win-rate comparison.}
\label{fig:bounded-ablation}
\end{figure*}

\textbf{Bounded context.} With capped top-$k$ retrieval and capped
item sizes, the configured prompt size is
$O(|\mathrm{sys}| + s_{\mathrm{thread}} + \sum_i k_i\cdot s_i)$.
The raw cross-decision transcript therefore does not scale with the
number of decisions. A transcript interface has worst-case
$\Omega(d\cdot \bar{s})$ growth for $d$ decisions, raising
per-decision token cost as a run lengthens \citep{lumer-cache-2026}. Figure~\ref{fig:bounded-ablation}a--b reports a per-cell linearity audit
($N{=}2$ runs per cell, $10$ runs total) of the released fixed-$A_0$ runs.

\textbf{Ablatable layers.} Because context reaches the model through
named slots, we can switch prompt strictness, rule retrieval,
episodes, and strategic skills on or off independently. This is the
main experimental advantage of the contract: the fixed-$A_0$ matrix
can ask which layer changes behavior, not only whether a larger
prompt helps.

\subsection{Five typed knowledge layers}
\label{sec:arch:layers}

The compose operator uses five substrates, separated by mutability
and role. $L_1$ \emph{operator prompts} contain immutable role and
protocol templates for each state type. $L_2$ \emph{state-typed
prompts} provide immutable schemas for combat, deckbuilding, map,
event, and intermission decisions, including legal action formats.
$L_3$ \emph{game knowledge} stores enumerable rule data---cards,
relics, events, enemies, and intents---refreshed by patch. $L_4$
\emph{episodic memory} stores postrun summaries (character $\times$
ascension $\times$ act $\times$ enemy class) --- \emph{case-based
recall}. $L_5$ \emph{skill library} stores triggered strategic
guides distilled from logs --- \emph{general scenario-class tactics}
indexed by trigger conditions for retrieval across recurring state
classes. Each $L_5$
guide has an explicit trigger, a prose policy, and a four-level
write gate.

Raw \stspad{} logs are not used as similarity RAG. In this game,
nearby-looking states can have very different strategic meanings
because of card order, relic combinations, and route history. The
agent therefore retrieves summaries and triggered guides rather than
nearest-neighbor log snippets.

\subsection{Routing and combat truncation}
\label{sec:arch:routing}

A dispatcher routes decisions to four model tiers: \emph{fast} for
trivial combat plans, \emph{strategic} for ordinary decisions,
\emph{analysis} for postrun memory extraction, and \emph{evolution}
for skill distillation. Four static system prompts are cacheable;
per-run state is placed in the user message. Combat is the only
decision type with a local conversation object, and that object emits
at most three messages per round: \verb|combat_start|, \verb|ok|,
and the latest user state. Earlier rounds are summarized through the
typed state rather than appended. Together with fast-tier routing
and mechanical handlers, this yields a median of 67 strategic LLM
calls per run instead of one call for every in-game action.

\subsection{\texorpdfstring{$L_4$}{L4} episodic memory: role and ablation}
\label{sec:arch:l4}

$L_4$ is the episodic layer. It stores postrun summaries for later
retrieval and can also feed online skill evolution. In the fixed-$A_0$
matrix, \texttt{full-frozen} (with $L_4$) and \texttt{mode-a}
(without $L_4$) both win $6/10$ games. Thus the balanced $A_0$
comparison points to $L_5$ as the layer associated with the headline
lift. $L_4$ remains part of the longer-horizon substrate used in the
auto-mode streams, which attempt $A_6$--$A_8$.

\subsection{Skill discovery: distillation, write gate, and Mode B}
\label{sec:arch:l5}

$L_5$ is populated in two ways. \textbf{Mistake-driven discovery}
(\emph{self-evolve}) reads combat losses relative to per-enemy
baselines, runs a pre-write A/B check ($B{=}3$ resample, strict
$2/3$ plus zero-harmful), and then applies a four-level write gate:
cosine, Jaccard, LLM judge, and optional reap. Most candidates are
rejected or merged rather than added as new skills. \textbf{Stub-template-filled
authoring} (\emph{Mode B}) fills five character-parametric templates
for combat, boss, deckbuilding, map, and intermission decisions,
under namespace isolation, a library lock, and warn-only validators.
Mode~B reaches the same $6/10$ fixed-$A_0$ point estimate as the
human-authored \texttt{mode-a} seed library (\S\ref{sec:results:modeb}),
so the evaluation can separate the existence of a skill layer from
the prose source used to populate it.


\section{Experimental Methodology}
\label{sec:methodology}

The evaluation is organized around three empirical questions. At a
single difficulty, which prompt and memory layers matter? If a frozen
$L_4{+}L_5$ stack is moved to another backbone, does it still help?
When postrun writing is allowed, how far does the agent climb on the
Ascension ladder?

\subsection{\texorpdfstring{The 5-condition decomposition at fixed $A_0$}{The 5-condition decomposition at fixed A0}}
\label{sec:methodology:fivecond}

\begin{figure*}[t]
  \centering
  \includegraphics[width=\linewidth]{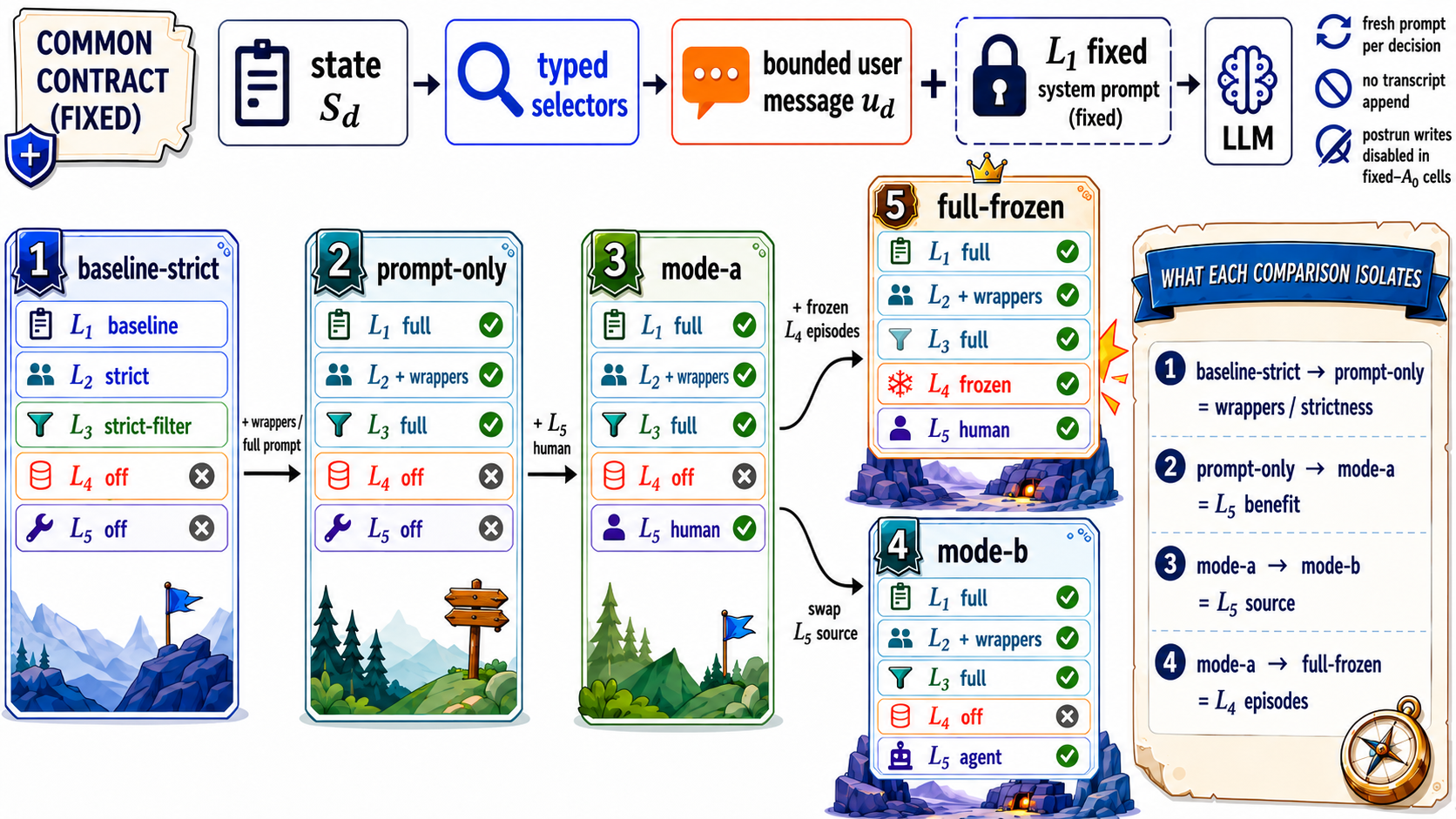}
  \caption{Fixed-$A_0$ ablation surface under the bounded-memory
  contract: five cells share a common contract and adjacent bars are
  organized by a named ablation axis. Not every neighboring pair is a
  single-mechanism isolation---e.g.\ baseline-strict to prompt-only
  changes the prompt-package/strictness group rather than one isolated
  mechanism. Frozen stores at SHA \texttt{1888a62}.}
  \label{fig:fivecond-surface}
\end{figure*}

The fixed-$A_0$ study is a five-cell decomposition, not a full
factorial grid (Figure~\ref{fig:fivecond-surface}).
\texttt{baseline-strict} uses the baseline prompt with no memory,
no skills, no strategic thread, and no combat-conversation wrapper;
it also applies the strict knowledge/hint filter. \texttt{prompt-only}
keeps the full prompt and conversation helpers but disables $L_4$ and
$L_5$. \texttt{mode-a} adds human-authored $L_5$ seed bodies.
\texttt{mode-b-frozen} replaces those bodies with stub-template-filled
$L_5$ bodies. \texttt{full-frozen} adds the frozen $L_4$ episodic
store to Mode~A. In all five cells, postrun and evolution writes are
disabled, anchoring the active stores at SHA \texttt{1888a62}.

\subsection{Cross-backbone probe and auto-mode ladder}
\label{sec:methodology:crossmodel}

The frozen $L_4{+}L_5$ stack was derived from Gemini~3.1~Pro
trajectories. To test how backbone-specific that stack is, we run
\texttt{baseline-strict} and \texttt{full-frozen} at $A_0$ on three
backbones: Qwen~3.6~27B, DeepSeek~V4~Pro, and Gemini~3.1~Pro. The
Qwen and DeepSeek probe adds $N{=}5$ completed games per
backbone-cell; the Gemini rows reuse the corresponding fixed-$A_0$
cells as anchors. We report this probe separately from the headline
ablation; Appendix~\ref{app:configs} lists the denominator rules.

The auto-mode ladder uses a different protocol. After a victory at
$A_n$, the next run attempts $A_{n+1}$; after a defeat, it retries
$A_n$. This stream measures the observed climb endpoint, whereas the
fixed-$A_0$ matrix measures reliability at one difficulty.

\subsection{Statistical protocol}
\label{sec:methodology:stats}

Cell-level win rates use Wilson 95\% confidence intervals
\citep{wilson1927}. Continuous scores (Eq.~\ref{eq:score}) use
5{,}000-bootstrap 95\% intervals \citep{efron1979}. The descriptive pooled scaffolded row uses an exact
Clopper--Pearson interval and is labeled as such. For the headline
fixed-$A_0$ table, we take the first ten completed games per
condition by start time, giving a balanced 50-game comparison. For
context, recent LLM-agent game evaluations typically report
3--25 episodes per condition---e.g., $3$ trials per cell in Voyager
\citep{voyager2023} and 10--25 seeds per task in BALROG
\citep{balrog2024}---so the present balanced $5\!\times\!10$ subset
sits at or above the comparable range while keeping completed-run
denominators interpretable. Completed games are the denominator
throughout. Additional completed games in the archive remain in
their diagnostic streams and are not pooled into the fixed-$A_0$
estimate. Score-based qualitative comparisons were also checked
after perturbing the $52/3$ coefficient by $\pm 10\%$; win-rate
intervals are unaffected.

\subsection{Reproducibility and release contents}
\label{sec:methodology:repro}

The public release includes the completed-run archive, condition
tags, analysis scripts, frozen memory/skill snapshots, and
representative prompt records needed to recompute the reported win
rates \citep{deal-checklist-2025,liu-versioning-2025}. The same
archive defines the natural next matched experiment: an
accumulating-context condition implemented in the same codebase with
the same run protocol and scoring scripts. Scope limits that affect
interpretation are noted where they arise and collected in the
Limitations section.

\section{Results}
\label{sec:results}

\subsection{Public difficulty calibration}
\label{sec:results:cross}

External rows calibrate difficulty, not causal attribution. The
May~2026 AGI-Eval snapshot reports zero listed $A_0$ victories
across five frontier-model configurations
\citep{agi-eval-spire-2026} (max defeat floor 33), and Mega Crit
reports a player-side $A_0$ win rate of $16\%$ across 240M community
runs \citep{megacrit-neowsletter-may2026}. Under our own harness,
\texttt{baseline-strict} wins $3/10$ runs (Wilson 95\% CI
$[10.8,60.3]$; mean score 70.4), placing the task in a hard but
non-saturated regime. Public rows use different interfaces, prompt
budgets, and decoding setups, so they are not matched baselines; the
within-harness ablation in \S\ref{sec:results:scaffold} establishes
the role of typed retrieval.

\subsection{Within-harness ablation}
\label{sec:results:scaffold}

Table~\ref{tab:fivecond} reports the balanced fixed-$A_0$ subset: ten
completed games per condition, the active $L_5/L_4$ layers, and the
mean derived score. The comparison isolates prompt strictness,
triggered skills, and episodic memory inside one codebase.

\begin{table}[!t]
\centering\footnotesize
\caption{Fixed-$A_0$ ablation ($N{=}10$/cell). Wilson 95\% CIs $[11,60]$/$[17,69]$/$[31,83]$ for $3/10$/$4/10$/$6/10$.}
\label{tab:fivecond}
\setlength{\tabcolsep}{3.5pt}
\renewcommand{\arraystretch}{1.0}
\begin{tabular*}{\columnwidth}{@{\extracolsep{\fill}}lccrr@{}}
\toprule
Cell & $L_5$ & $L_4$ & Win & Score \\
\midrule
No scaffold      & -- & -- & 3/10 & 70.4 \\
Prompt only      & -- & -- & 4/10 & 69.6 \\
\midrule
Hand skills      & A  & -- & \textbf{6/10} & \textbf{85.5} \\
Template skills  & B  & -- & \textbf{6/10} & 83.3 \\
Skills+episodes  & A  & $\checkmark$ & \textbf{6/10} & 82.1 \\
\bottomrule
\end{tabular*}
\end{table}

The largest separation is between no-scaffold and skill-scaffolded
rows. The two no-scaffold cells win $3/10$ and $4/10$ games; all
three $L_5$ cells win $6/10$. Writing the layer-attributable
difference as
\begin{equation}
\Delta_{L_\ell} = \widehat p_{\text{with-}\ell} - \widehat p_{\text{without-}\ell},
\label{eq:lift}
\end{equation}
Table~\ref{tab:fivecond} gives $\Delta_{\text{prompt}}{=}{+}1/10$
(strictness, wrappers) and $\Delta_{L_5}{=}{+}2/10$ at the same prompt
setup. \emph{At $N{=}10$ this difference is not statistically
significant}: a Fisher exact test on $3/10$ vs.\ $6/10$ gives
$p\approx0.37$, and pooling all scaffolded vs.\ unscaffolded cells
($18/30$ vs.\ $7/20$) gives $p\approx0.148$; the Wilson 95\% CIs
\citep{wilson1927} overlap. We therefore read $L_5$ as the layer with
the largest observed difference in the balanced matrix --- a
directional result, not a fine ranking among the three scaffolded
variants and not a significance claim. Establishing whether the
bounded contract itself outperforms a matched accumulating-context
design would require the controlled comparison we leave to future
work (Limitations).

\subsection{Auto-mode ascension ladder: endpoint evidence}
\label{sec:results:auto}

\begin{figure}[t]
\centering
\includegraphics[width=0.52\linewidth]{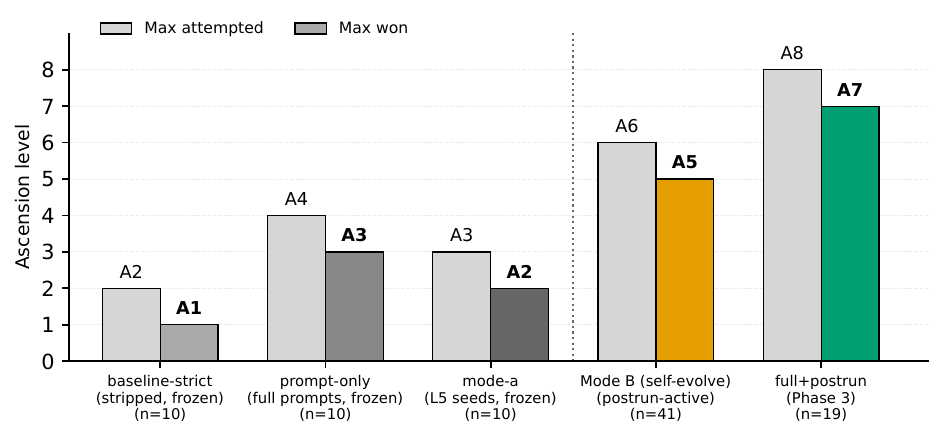}
\caption{Auto-mode ascension ladder: per-stream highest attempted
ascension (endpoint, not win-rate).}
\label{fig:ascend}
\end{figure}

Runs with postrun-active memory attempt $A_6$--$A_8$, while
no-postrun streams stop at $A_2$--$A_4$
(Figure~\ref{fig:ascend}). The ladder therefore complements the
fixed-$A_0$ matrix: one stream isolates component lift at a fixed
difficulty, and the other shows the highest difficulty reached
when stores can be updated after runs.

\subsection{Template skills, transfer, and episodes}
\label{sec:results:modeb}

\paragraph{Mode B without hand-authored skill prose.}
\texttt{mode-b-frozen} matches the hand-authored \texttt{mode-a} 60\%
$A_0$ win estimate (score-diff CI $[-18.6,+24.8]$): template-filled
skills are competitive with the seed library inside the same skill
interface.

\paragraph{Frozen skills are backbone-sensitive.}
Table~\ref{tab:crossbackbone} shows the same Gemini-trained
\texttt{full-frozen} $L_4{+}L_5$ stack transferred to two
non-training backbones. The Gemini-trained stack lifts Qwen's
mean score ($+84.5\%$) but reduces DeepSeek's ($-18.1\%$); Qwen and
DeepSeek wins remain $0/5$. Transfer is therefore an empirical
property of the stack, not a premise.

\begin{table}[!t]
\centering\scriptsize
\caption{Cross-backbone transfer of the Gemini-trained
$L_4{+}L_5$ stack ($N{=}5$/cell, Gemini $N{=}10$). Full-frozen 95\%
score CIs $[13.8,41.9]$/$[21.7,45.9]$/$[63.1,96.5]$ for
Qwen/DeepSeek/Gemini; Qwen and DeepSeek wins $=0/5$ in both columns,
so $\Delta\%$ is a score-only signal. $^{\dagger}$Gemini
floor-48 endpoints include victories.}
\label{tab:crossbackbone}
\setlength{\tabcolsep}{3pt}
\renewcommand{\arraystretch}{1.0}
\begin{tabular*}{\columnwidth}{@{\extracolsep{\fill}}lcccc@{}}
\toprule
Backbone & Wins & Score & $\Delta$\% & Floor \\
\midrule
Qwen 3.6-27B    & 0/5$\to$0/5            & $14.6\!\to\!\textbf{26.9}$ & $+84.5$ & $17\!\to\!\textbf{33}$ \\
DeepSeek V4-Pro & 0/5$\to$0/5            & $41.3\!\to\!33.8$          & $-18.1$ & $37\!\to\!33$ \\
Gemini 3.1-Pro  & 3/10$\to$\textbf{6/10} & $70.4\!\to\!\textbf{82.1}$ & $+16.6$ & $48\!\to\!48^{\dagger}$ \\
\bottomrule
\end{tabular*}
\end{table}

\paragraph{$L_4$ at $A_0$ is saturated.}
\label{sec:results:l4}
\texttt{mode-a} (no $L_4$) and \texttt{full-frozen} (with $L_4$)
produce the same win point estimate (score-diff CI $[-21.7,+14.9]$);
$L_4$ still serves the longer-horizon substrate in the ladder
streams (\S\ref{sec:results:auto}).


\section{Comparison with Open-Source Accumulating-Context Agents}
\label{sec:competitors}

The submitted version compared against external calibration anchors only. Here we add a
direct, same-testbed \emph{operational} comparison against the two open-source StS2 agents
that could be run faithfully end-to-end: \textbf{STS2MCP}~\citep{sts2mcp-2026} and
\textbf{CharTyr}~\citep{chartyr-sts2-2026}. Both follow the dominant agent-loop design our
contract argues against: a single accumulating chat transcript, re-sent (and grown) on
every decision. We stress at the outset that this is a comparison of \emph{shipped systems},
not a controlled ablation of the memory contract: the competitors differ from our agent in
game patch, routing, thinking effort, decision batching, and prompt cadence as well as in
the contract, so the gaps below characterize the current state of practice rather than
isolating boundedness as the cause.

\paragraph{Faithful replication.}
Each competitor runs its \emph{author-intended} configuration: the author's own mod build
(minimal load-compatibility patches only; zero agent-logic changes), the author's own MCP
server, and the author's own skill/strategy documents as the system prompt, driven through
their tool interface. A leak audit over the captured requests confirms no content from our
project enters their context. Exactly one mod is loaded at a time, and every game is a
fresh Silent $A_0$ run on the same machine and the same v0.103.x line of the game: our
cells ran on v0.103.1; a minor game patch (v0.103.3, 2026-05-30) landed between the two
batches, so competitor runs used v0.103.3 (their mods compile against the v0.103.1-pinned
reference and load cleanly on v0.103.3; our own stack re-verified on v0.103.3). All strategic
decisions for every agent run on \texttt{gemini-3.1-pro-preview}; our agent additionally
routes trivial decisions to a flash-lite fast tier and sets explicit thinking effort,
while competitors run at the provider default with no thinking parameter (their intended
setup). The denominator is completed games --- harness failures are re-run, never counted
as losses --- and every run ended in a natural in-game terminal, none at the decision cap.
All raw prompts, responses, token usage, and per-step game state are released
(released with the data archive).

\begin{figure}[t]
  \centering
  \includegraphics[width=\linewidth]{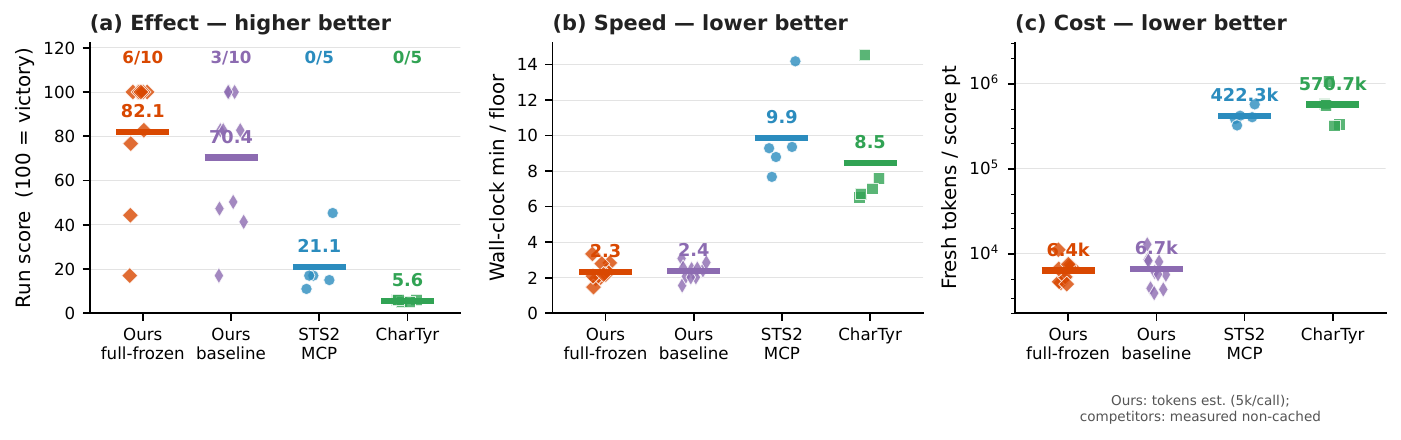}
  \caption{Competitor comparison at $A_0$ (The Silent), per run; horizontal bars are
  per-cell means. \textbf{(a) Effect}: run score ($s=100$ if victory else
  $\mathrm{floor}+(52/3)\cdot\mathrm{bosses}$), with win counts above each column.
  \textbf{(b) Speed}: wall-clock minutes per floor reached (96\% of competitor wall-clock
  is provider-reported LLM latency; fixed inter-action delays slightly favor competitors,
  0.5s vs.\ our 0.6s; our durations exclude postrun). \textbf{(c) Cost}: fresh (non-cached)
  LLM tokens per score point (log). Competitor usage is exact provider-reported tokens with
  cache hits removed (90\%\,/\,82\% of their prompt tokens were cached); ours follows the
  paper's Fig.~3 convention ($\sim$5k strategic user-message tokens $\times$
  \texttt{llm\_calls}), excluding the cached system prefix, completions, retries, and
  fast-tier calls. \texttt{full-frozen}/\texttt{baseline-strict} are the Table~2 cells
  ($N{=}10$, frozen stores at SHA \texttt{1888a62}); competitors $N{=}5$. Under the
  intentionally absurd upper bound pricing \emph{every} recorded action as a full strategic
  call, our cells move to 55k\,/\,58k tokens per point --- the gap remains $\geq 7\times$.}
  \label{fig:cmp-dashboard}
\end{figure}

\paragraph{Effect.}
Figure~\ref{fig:cmp-dashboard}(a) shows run scores ($s = 100$ if victory else
$\mathrm{floor} + (52/3)\cdot\mathrm{bosses}$, \S\ref{sec:testbed}). Both accumulating
agents collapse at $A_0$: STS2MCP wins $0/5$ (mean floor 17.6; one Act-1 boss cleared
across five runs), CharTyr wins $0/5$ (mean floor 5.6; its frequent
\texttt{invalid\_action} interface errors compound into early deaths --- a property of the
agent under test, faithfully reproduced\footnote{CharTyr outcomes for runs 2--5 are
inferred from terminal floors 5--6 plus the absence of any victory flag anywhere in the
captures; run 1's defeat is confirmed by its captured structured
\texttt{game\_over.is\_victory=false}. All captures are released for audit.}). Our
\texttt{full-frozen} cell wins $6/10$ (mean score 82.1) and even \texttt{baseline-strict}
--- our harness with the bounded contract but no learned stores --- wins $3/10$ (70.4),
so the gap is not explained by harness quality alone.

\begin{figure}[t]
  \centering
  \includegraphics[width=\linewidth]{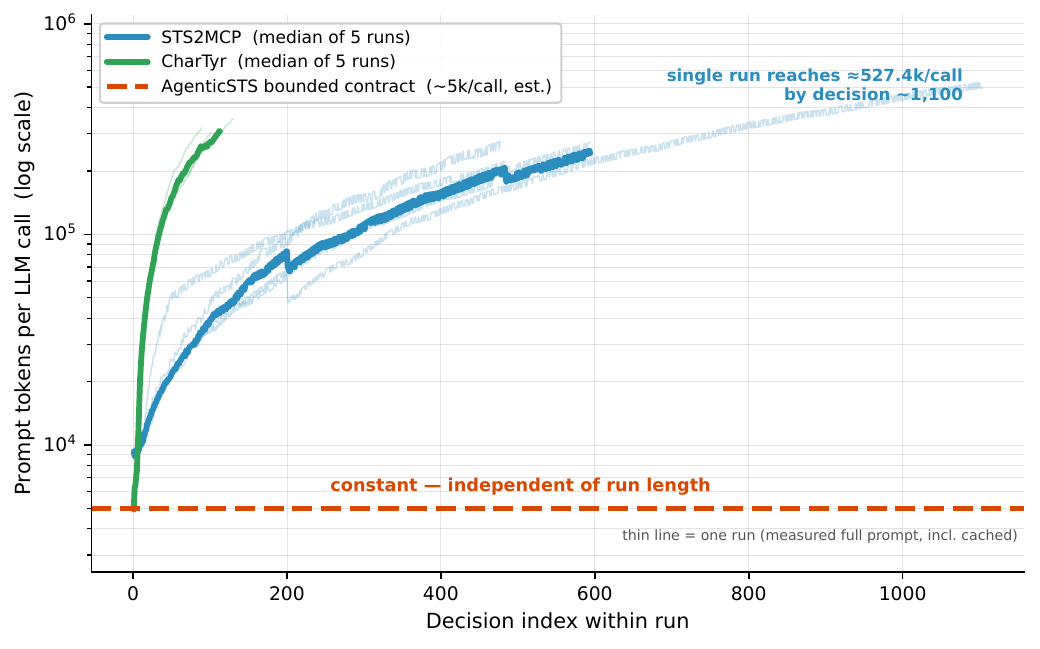}
  \caption{The mechanism: per-call prompt size over a run. Thin lines are individual runs
  (measured full prompts, including cached tokens --- caching changes billing, not what the
  model attends to); bold lines are per-competitor medians. The worst single STS2MCP run
  reaches $\sim$500k tokens per call by decision $\sim$1100. The dashed line is our bounded
  contract's strategic user-message median ($\sim$5k, estimate; constant cached system prefix
  excluded; $x$-extent not comparable --- our runs make $\sim$100 strategic calls).}
  \label{fig:cmp-growth}
\end{figure}

\begin{figure}[t]
  \centering
  \includegraphics[width=\linewidth]{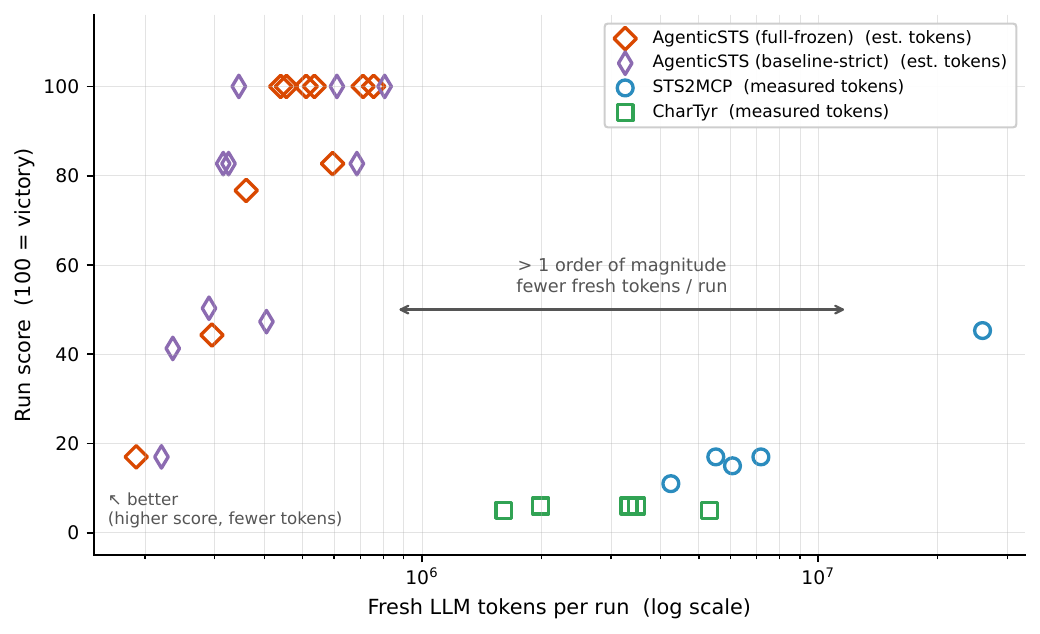}
  \caption{Cost--effect frontier: run score vs.\ fresh LLM tokens per run (log). Hollow
  diamonds: our cells (tokens estimated, 5k/call convention); circles/squares: measured
  competitor usage. The empty band between the clusters spans more than an order of
  magnitude.}
  \label{fig:cmp-pareto}
\end{figure}

\paragraph{Speed and cost.}
Figure~\ref{fig:cmp-dashboard}(b,c) and Figures~\ref{fig:cmp-growth}--\ref{fig:cmp-pareto}
quantify the operational gap.
Per floor reached, the accumulating agents need $\sim$4$\times$ the wall clock
(9.9\,/\,8.5 vs.\ 2.3 minutes; 96\% of their wall-clock is provider-reported LLM latency,
so the gap is not harness pacing). Per score point, they spend $66$--$90\times$ more
\emph{fresh} (non-cached) LLM tokens; under raw ingested context the multiplier exceeds
$450\times$, and even pricing every recorded action of ours as a full strategic call
(an intentionally absurd upper bound) leaves $\geq 7\times$. Part of this gap is decision
batching --- one strategic call drives multiple actions in our agent, while the
competitors call the LLM once per action by design --- which we report as a property of
the memory-architecture package rather than of the backbone.
Figure~\ref{fig:cmp-growth} shows the mechanism: their per-call prompt grows from
$\sim$9k to $500$k tokens within a single run (trimming caps message \emph{count}, but
late-game states grow each message), while the bounded contract holds the strategic user
message flat at a $\sim$5k median.

\paragraph{What this comparison does and does not show.}
It does not show that accumulating context can never win StS2 --- both competitors are
community projects, not tuned baselines, and CharTyr's losses are partly interface
errors. It does show that the two publicly available transcript-accumulating agents,
run faithfully on the same backbone, game line, character, and ascension, with the
denominator rules of our own evaluation, fall far below even our no-store bounded
baseline while consuming one to two orders of magnitude more tokens per point of
progress. A matched same-codebase accumulating-context cell would isolate the
contract variable itself; we leave that controlled comparison to follow-up work,
and this release is organized to support it. The present section establishes the
external state of practice under disclosed operational differences.

\section{Discussion}
\label{sec:discussion}

\paragraph{Interpretation and scope.}
The main lesson is that the memory interface can be made into an
object of evaluation rather than left as a prompting convention. In
a closed-rule game with text-readable state, the bounded typed
contract supports fixed-$A_0$ wins, locates the largest
within-harness difference at the $L_5$ skill layer (directional at
our sample size), and keeps fixed-difficulty performance separate
from ladder endpoints. The release is meant to
make the next comparison easier: an accumulating-context variant can
be added in the same codebase, with the same condition tags, frozen
stores, prompt records, and scoring scripts, rather than inferred
from public runs that use different harnesses.

\paragraph{Implications.}
Two observations suggest broader applicability. First, separating
memory into typed slots makes attribution tractable: gains can be
traced to a specific layer rather than to ``more context.'' Second,
the bounded contract decouples interface design from accumulating
state, making the same evaluation surface portable to non-game
agentic tasks with comparable closed-rule structure.

\paragraph{Implications for loop engineering.}
The bounded contract is a concrete, measurable design point for the memory
stage of closed-rule, turn-based agent loops: per-decision typed retrieval
keeps the online context bounded regardless of run length, typed stores make
memory updates auditable, and postrun writes expose learning as explicit
artifacts rather than opaque transcript growth. Whether this pattern is the
right default for open-ended production loops is untested here; what we provide
is a reproducible testbed and archive for measuring a memory-layer change under
fixed game, denominators, and scoring---an empirical complement to the largely
qualitative loop-engineering guidance now common in
practice~\citep{anthropic-context-2025,memory-survey-2025}.

\section{Conclusion}
\label{sec:conclusion}

We present a long-horizon LLM-agent resource where memory is bounded,
typed, and ablatable at decision time. In the fixed-$A_0$ matrix the
largest observed difference appears when triggered $L_5$ skills are
enabled, though at this sample size it is directional rather than
statistically decisive; the $298$-trajectory archive supplies logs,
frozen stores, prompts, and scripts. By keeping game, denominators,
stores, and prompts aligned, the release lets future work ask whether
a new memory interface changes decisions rather than whether it
benefited from a different evaluation scaffold. Our evidence supports
a narrower conclusion: explicit, typed memory contracts make
long-horizon agent behavior easier to audit, reproduce, and ablate.
Whether a bounded contract outperforms a matched accumulating-context
design remains an open question for a controlled follow-up.

\section*{Limitations}
\label{sec:limitations}

\paragraph{Sample size and statistical inference.}
The headline fixed-$A_0$ result uses a balanced 50-game subset,
which is at or above the per-condition sample size typical for
recent LLM-agent game benchmarks
\citep{voyager2023,balrog2024,lmgame-bench-2025}. The
cross-backbone and ladder streams are smaller diagnostic streams,
reported separately rather than pooled into the headline table.
Wilson intervals at this sample size are read at the
scaffolded-versus-unscaffolded level; finer-grained equivalence
tests among scaffolded variants and smooth backbone-transfer
curves are natural extensions that the released archive supports.

\paragraph{Same-codebase accumulating-context variant.}
The release is organized so that an accumulating-context cell
sharing our codebase, condition tags, scoring scripts, and frozen
stores can be added as one further row in the same matrix. This is
the cleanest direct comparison to the bounded contract, and the
artifact provides the run protocol, prompt records, and analysis
scripts needed to run it.

\paragraph{Single character and game-version coverage.}
The headline runs target one playable character, Silent, chosen to
keep the typed substrate ($L_3$ enumerable game knowledge,
$L_4{+}L_5$ stores) self-consistent in the present submission.
Released trajectories carry game-version tags so that
version-stratified re-analysis is possible from the archive, and
cross-character runs follow the same harness once $L_3/L_4/L_5$
are repopulated for the new character.

\paragraph{External player and ecosystem references.}
Mega Crit, Spiracle, and AGI-Eval enter the paper only as
difficulty and ecosystem context, with cached snapshots released
alongside the artifact. Matched human inference would require a
separately designed user study with controlled denominators, which
is outside the present resource.

\paragraph{Architectural scope.}
The evaluation is training-free and single-game. The bounded
contract is tuned for \emph{turn-based decision} settings like
\stspad{}; continuous or streaming control loops, visual input,
multi-agent play, online human correction, model-internal
fine-tuning, and cross-game transfer are deliberate non-targets of
this release. Stub templates and expert seed skills are
author-curated; Mode~B measures within-interface template filling,
which we report as one operating point rather than as fully
autonomous skill invention.


\bibliographystyle{abbrv}
\bibliography{refs}

\beginappendix

\section{Evaluation archive and aggregation rule}
\label{app:configs}

The archive is organized so that each reported win rate can be
recomputed from completed-run records. It includes one record per
trajectory, the condition tags used to select cells, frozen $L_4{+}L_5$
snapshots for the within-codebase ablation, representative
decision-time and postrun prompt records, and scripts for Wilson 95\%
win-rate intervals and bootstrap 95\% score intervals. A cell's win
rate is simply victories divided by completed games in that cell. We
keep the fixed-$A_0$, cross-backbone, ladder, and full-archive
streams separate. Table~\ref{tab:evidence-streams} gives the
denominator rule for each stream.

\begin{table}[H]
\centering\footnotesize
\caption{Denominator map. Only the balanced fixed-$A_0$ subset enters the headline ablation (Table~\ref{tab:fivecond}); streams are never pooled.}
\label{tab:evidence-streams}
\setlength{\tabcolsep}{4pt}
\renewcommand{\arraystretch}{1.08}
\begin{tabular*}{\columnwidth}{@{\extracolsep{\fill}}llll@{}}
\toprule
Stream & $N$ & Metric & Role \\
\midrule
Fixed-$A_0$ & 50 $(5\!\times\!10)$ & win+score & headline \\
Backbone & 5/cell & score shift & diagnostic \\
Ladder & unequal & max $A$ & endpoint \\
Archive & 298 & tags+scripts & audit \\
\bottomrule
\end{tabular*}
\end{table}
\FloatBarrier

\paragraph{Score-formula audit.}
The derived score (Eq.~\ref{eq:score}) uses $\mathrm{bosses}=0$
when floor $<18$, $1$ when floor $<34$, $2$ otherwise, and $3$
for victories. Table~\ref{tab:cell-floor-boss} lets readers
reproduce the paper-reported means.

\paragraph{Confidence intervals.}
For a cell with $w$ wins out of $n$ completed runs, the Wilson 95\%
interval \citep{wilson1927} on the win rate $\hat p = w/n$ is
\begin{equation}
\frac{\hat p + \tfrac{z^2}{2n} \pm z\sqrt{\tfrac{\hat p(1-\hat p)}{n} + \tfrac{z^2}{4n^2}}}{1 + z^2/n},
\quad z=1.96.
\label{eq:wilson}
\end{equation}
Score intervals use the percentile bootstrap \citep{efron1979}: 5{,}000
resamples with replacement from the cell's $n$ run-level scores,
reporting the empirical $[2.5,97.5]$ percentiles. The descriptive
pooled scaffolded row uses an exact Clopper--Pearson interval and is
labeled as such.

\begin{table}[H]
\centering\footnotesize
\caption{Per-cell floor and boss-clear counts that reproduce the
mean scores reported in Table~\ref{tab:fivecond}.}
\label{tab:cell-floor-boss}
\setlength{\tabcolsep}{4pt}
\renewcommand{\arraystretch}{1.05}
\begin{tabular*}{\columnwidth}{@{\extracolsep{\fill}}lccccc@{}}
\toprule
Cell & $N$ & Wins & $\overline{\mathrm{floor}}$ & $\overline{\mathrm{bosses}}$ & $\overline{\mathrm{score}}$ \\
\midrule
baseline-strict & 10 & 3 & 39.2 & 1.80 & 70.40 \\
prompt-only     & 10 & 4 & 38.4 & 1.80 & 69.60 \\
mode-a          & 10 & 6 & 43.9 & 2.40 & 85.50 \\
mode-b-frozen   & 10 & 6 & 43.4 & 2.30 & 83.27 \\
full-frozen     & 10 & 6 & 42.2 & 2.30 & 82.07 \\
\bottomrule
\end{tabular*}
\end{table}
\FloatBarrier

\section{Memory contract: five typed layers}
\label{app:memory-contract}

Table~\ref{tab:memory-contract} gives a compact legend for the five
stores used in the paper. The useful distinction is mutability:
$L_1$ and $L_2$ are fixed, $L_3$ can be filtered, and $L_4/L_5$ can be
disabled, frozen, or made writable depending on the evidence stream.

\begin{table}[H]
\centering\footnotesize
\caption{Five-layer memory contract. Each decision receives typed slices; raw cross-decision transcript is not appended.}
\label{tab:memory-contract}
\setlength{\tabcolsep}{4pt}
\renewcommand{\arraystretch}{1.08}
\begin{tabular*}{\columnwidth}{@{\extracolsep{\fill}}cllll@{}}
\toprule
Layer & Store & Key & Write & Ablation \\
\midrule
$L_1$ & protocol & state & fixed & always \\
$L_2$ & schema & decision & fixed & always \\
$L_3$ & rules & entities & refresh & filter \\
$L_4$ & episodes & char/$A$/act & postrun & off/on \\
$L_5$ & skills & trigger & gated & off/A/B \\
\bottomrule
\end{tabular*}
\end{table}
\FloatBarrier

\section{Concrete prompt and learning examples}
\label{app:artifacts}

\paragraph{Decision-time composed prompt.}
At decision time, the agent assembles the user prompt in a fixed
typed order: fired $L_5$ skills, $L_4$ episodes, $L_3$ game facts,
the $L_2$ typed state prompt, and a schema hint listing valid
actions. The order is shared across non-combat decision states. Combat
is the only intra-fight stateful container, and its emitted message
list per round is bounded to three items: combat start, an
acknowledgement, and the latest user state. Earlier rounds are not
copied into the transcript; the next round is prompted from a newly
composed user message. A shortened floor-7 card-reward excerpt shows
the typed layers:

\begin{Verbatim}[fontsize=\footnotesize]
## Expert Knowledge (retrieved skills)
**Silent - Draft and Shop Rules** (seed)
Early rewards must solve damage first,
then block, draw, and energy. ...

## Card-Specific Insights
- dodge and roll: delayed block plus Dexterity.
- slice: 0-cost transitional damage.

## Game Knowledge / Card Mechanics
- DodgeAndRoll: Block Next Turn; upgrade +2.
- Slice: Upgrade: Damage +3

## Card Reward
HP: 57/57 (100%) | Gold: 55 | Act: 1 | Floor: 7

## Decision Format (card_reward_action)
Valid actions: choose_reward_card
or choose_reward_alternative
\end{Verbatim}

\paragraph{Representative learned skill.}
One learned skill concerns a boss mechanic that exhausts attacks and
skills on specific turns (turns 2, 5, 8, and 11). The stored rule
instructs the agent to preserve core scaling cards on those turns and
to prefer powers or natively exhausting cards. The example shows that
$L_5$ stores state-conditioned tactical rules with Boolean triggers,
not generic advice or similarity-retrieved raw logs. Full prompt and
learned-skill records are included in the artifact archive.

\section{External calibration and source boundaries}
\label{app:difficulty-calibration}

External \stspad{} statistics enter the paper as difficulty and
ecosystem context. Cached snapshots of each source are released
with the artifact archive so that follow-up work can reproduce
the calibration rows under identical inputs.

\begin{table}[H]
\centering\footnotesize
\caption{External numbers calibrate difficulty and ecosystem context only.}
\label{tab:difficulty-calibration}
\setlength{\tabcolsep}{4pt}
\renewcommand{\arraystretch}{1.08}
\begin{tabular*}{\columnwidth}{@{\extracolsep{\fill}}lll@{}}
\toprule
Source & Role & Not used as \\
\midrule
AGI-Eval & LLM yardstick & causal baseline \\
Mega Crit $A_0$ & difficulty anchor & matched human test \\
Spiracle & community context & population rate \\
\bottomrule
\end{tabular*}
\end{table}
\FloatBarrier



\clearpage
\section{Full decision-time prompt exhibits}
\label{app:full-prompt-exhibits}

The exhibits below print the model-facing prompt assembled from the
typed substrates. Text inside each box is verbatim system or user
prompt content; the box colors and titles are reader annotations for
$L_1$ protocol/schema, $L_2$ current state and action space, $L_3$
rules and mechanics, $L_4$ episodic notes, and $L_5$ strategic
skills.

\begin{figure*}[t]
\centering
\includegraphics[width=.48\textwidth]{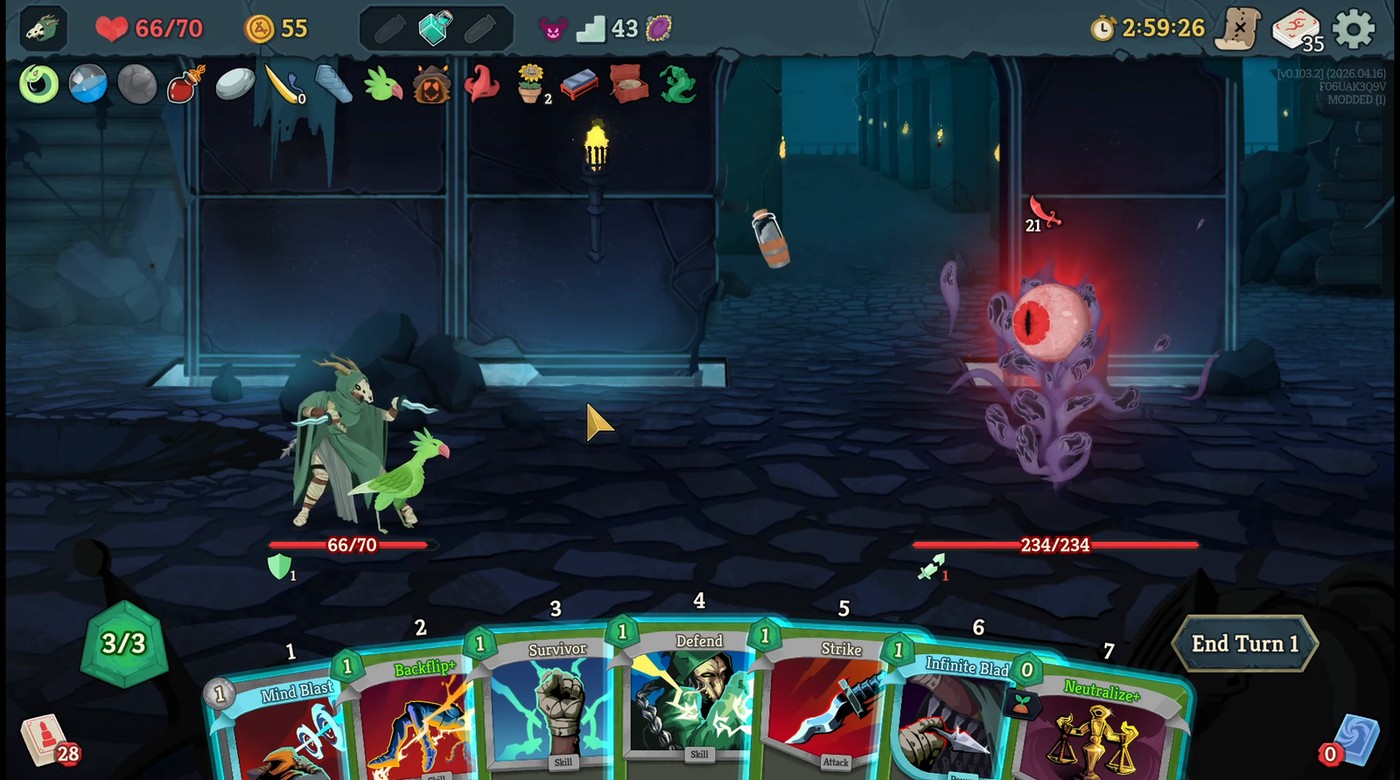}\hfill
\includegraphics[width=.48\textwidth]{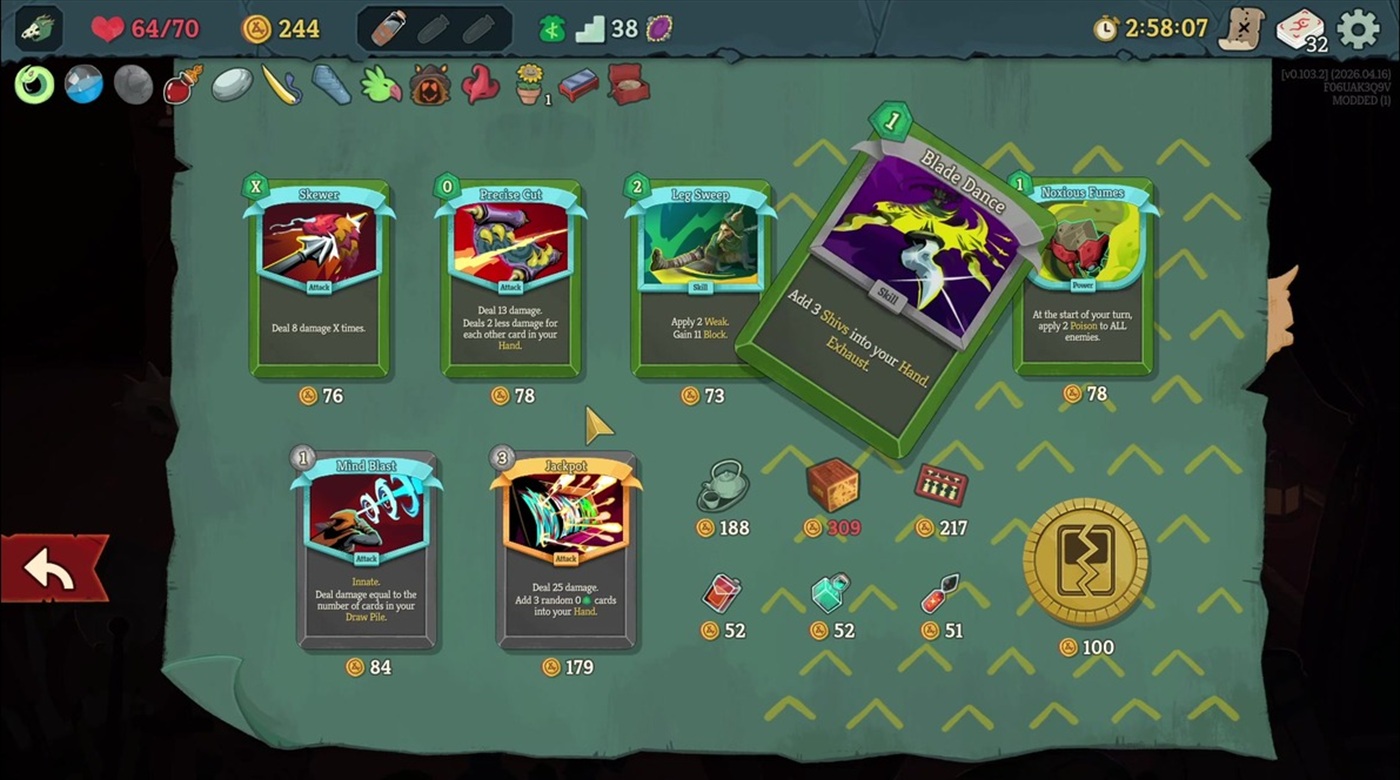}
\caption{Decision states used in the prompt exhibits: combat planning (left) and shop planning (right).}
\label{fig:prompt-exhibit-states}
\end{figure*}

\subsection{Combat decision}
\label{app:full-prompt-combat}

The combat exhibit is the longest example: it includes the shared
system instruction, setup context, retrieved skills, episodic notes,
game rules, and the current round state.

\paragraph{System prompt.}
\LoneFile{Role}{appendix_prompts/combat_system_01_L1_preamble.txt}
\LoneFile{Output schema}{appendix_prompts/combat_system_02_L1_output_format.txt}
\LthreeFile{Combat rules}{appendix_prompts/combat_system_03_L3_core_combat_rules.txt}
\LfiveFile{HP policy}{appendix_prompts/combat_system_04_L5_hp_conservation.txt}

\paragraph{User prompt: setup and retrieved context.}
\LtwoFile{Combat start}{appendix_prompts/combat_user_setup_01_L2_combat_start.txt}
\LtwoFile{Deck}{appendix_prompts/combat_user_setup_02_L2_current_deck_35_cards.txt}
\LtwoFile{Relics}{appendix_prompts/combat_user_setup_03_L2_relics_14.txt}
\LfourFile{Strategic thread}{appendix_prompts/combat_user_setup_04_L4_strategic_thread.txt}
\LfiveFile{Retrieved skills}{appendix_prompts/combat_user_setup_05_L5_expert_knowledge_retrieved_skills.txt}
\LfourFile{Past experience}{appendix_prompts/combat_user_setup_06_L4_past_experience.txt}
\LfourFile{Past experience}{appendix_prompts/combat_user_setup_07_L4_past_experience.txt}
\LfourFile{Combat guide}{appendix_prompts/combat_user_setup_08_L4_combat_guide.txt}
\LtwoFile{Enemy patterns}{appendix_prompts/combat_user_setup_09_L2_enemy_patterns.txt}
\LfiveFile{Potion strategy}{appendix_prompts/combat_user_setup_10_L5_potion_strategy.txt}
\LfourFile{Card notes}{appendix_prompts/combat_user_setup_11_L4_card_notes_from_experience.txt}
\LthreeFile{Rules}{appendix_prompts/combat_user_setup_12_L3_combat_rules.txt}

\paragraph{User prompt: current combat state.}
\LtwoFile{Round state}{appendix_prompts/combat_user_state_01_L2_round_1_state.txt}
\LtwoFile{Enemies}{appendix_prompts/combat_user_state_02_L2_enemies.txt}
\LtwoFile{Relic counters}{appendix_prompts/combat_user_state_03_L2_relic_counters.txt}
\LtwoFile{Potions}{appendix_prompts/combat_user_state_04_L2_usable_potions.txt}
\LtwoFile{Piles}{appendix_prompts/combat_user_state_05_L2_piles.txt}
\LthreeFile{Active effects}{appendix_prompts/combat_user_state_06_L3_key_effects_active_this_combat.txt}
\LtwoFile{Hand}{appendix_prompts/combat_user_state_07_L2_hand_7_playable_7_total.txt}

\paragraph{Structured response.}
The model returned a structured combat plan: use Powdered Demise,
play Neutralize+, Mind Blast, Backflip+, Infinite Blades, Survivor,
and then end the turn.

\subsection{Shop-planning decision}
\label{app:full-prompt-shop}

The shop exhibit shows the same interface outside combat. To avoid
printing the shared role and generic JSON schema twice, it includes
only the shop-specific system additions and the shop user prompt.

\paragraph{Shop-specific system prompt.}
\LfiveFile{Deck philosophy}{appendix_prompts/shop_system_03_L5_card_deck_philosophy.txt}
\LfiveFile{Two-phase framework}{appendix_prompts/shop_system_04_L5_strategic_deckbuilding_the_two_phase_framework.txt}
\LfiveFile{Phase 1}{appendix_prompts/shop_system_05_L5_phase_1_foundation_no_engine_yet.txt}
\LfiveFile{Phase 2}{appendix_prompts/shop_system_06_L5_phase_2_commitment_engine_acquired.txt}
\LoneFile{Note schema}{appendix_prompts/shop_system_07_L1_output_strategic_note.txt}

\paragraph{User prompt.}
\LfiveFile{Retrieved skills}{appendix_prompts/shop_user_01_L5_expert_knowledge_retrieved_skills.txt}
\LfourFile{Deck insights}{appendix_prompts/shop_user_02_L4_deck_building_insights.txt}
\LfourFile{Card insights}{appendix_prompts/shop_user_03_L4_card_specific_insights.txt}
\LthreeFile{Game knowledge}{appendix_prompts/shop_user_04_L3_game_knowledge.txt}
\LthreeFile{Card mechanics}{appendix_prompts/shop_user_05_L3_card_mechanics_from_game_data.txt}
\LtwoFile{Shop state}{appendix_prompts/shop_user_06_L2_shop.txt}
\LtwoFile{Deck}{appendix_prompts/shop_user_07_L2_current_deck_32_cards.txt}
\LtwoFile{Relics}{appendix_prompts/shop_user_08_L2_relics_ring_of_the_snake_at_the_start_of_each_combat_draw_2_additional_cards_lar.txt}
\LfourFile{Relic synergies}{appendix_prompts/shop_user_09_L4_relic_synergies.txt}
\LfourFile{Gold budget}{appendix_prompts/shop_user_10_L4_gold_budget_analysis.txt}
\LfiveFile{Shop guide}{appendix_prompts/shop_user_11_L5_guide.txt}
\LfourFile{Card notes}{appendix_prompts/shop_user_12_L4_card_notes.txt}
\LoneFile{Task}{appendix_prompts/shop_user_13_L1_your_task.txt}
\LthreeFile{Keywords}{appendix_prompts/shop_user_14_L3_keyword_glossary.txt}
\LtwoFile{Items}{appendix_prompts/shop_user_15_L2_items_for_sale.txt}
\LoneFile{Shop-plan schema}{appendix_prompts/shop_user_16_L1_decision_format_shop_plan.txt}

\paragraph{Structured response.}
The model returned a shop plan that buys Blade Dance, Leg Sweep, Mind
Blast, and Swift Potion, while skipping the remaining affordable items
with item-level reasons.

\end{document}